\documentclass[lettersize,journal]{IEEEtran}
\usepackage{amsmath,amsfonts}
\usepackage{algorithmic}
\usepackage{algorithm}
\usepackage{array}
\usepackage[caption=false,font=normalsize,labelfont=sf,textfont=sf]{subfig}
\usepackage{textcomp}
\usepackage{stfloats}
\usepackage{url}
\usepackage{verbatim}
\usepackage{graphicx}
\usepackage{cite}
\usepackage[colorlinks=true,
            linkcolor=blue,
            citecolor=blue,
            urlcolor=blue]{hyperref}  % 用于处理链接

\usepackage{booktabs}
\graphicspath{{Figures/}}

\usepackage{multirow}
\usepackage{threeparttable}
\usepackage{setspace}
\usepackage{makecell}
\begin{document}

\title{Design of an Underwater Vehicle With Orientation-Adjustable Thrusters and Adaptive Tracking Control}

\author{Yifei Wang, Shihan Kong, Zhanhua Xin, Kaiwei Zhu, and Junzhi Yu,~\IEEEmembership{Fellow,~IEEE}
        % <-this % stops a space
\thanks{
Received 19 August 2025; revised 24 December 2025; accepted 5 March 2026. \emph{(Yifei Wang and Shihan Kong contributed equally to this work.) (Corresponding author: Junzhi Yu.)}}	
		
\thanks{Yifei Wang, Shihan Kong, Zhanhua Xin, Kaiwei Zhu, and Junzhi Yu are with the State Key Laboratory for Turbulence and Complex Systems, School of Advanced Manufacturing and Robotics, Peking University, Beijing 100871, China (e-mail: yifei@stu.pku.edu.cn; kongshihan@pku.edu.cn; xinzhanhua@stu.pku.edu.cn; kw{\_}zhu@stu.pku.edu.cn; junzhi.yu@ia.ac.cn).}
}



\maketitle
\begin{abstract}
Unmanned underwater vehicles (UUVs) are essential for marine exploration and research. However, conventional designs often struggle with limited maneuverability in complex and dynamic underwater environments. This paper presents an orientation-adjustable thruster UUV (OAT-UUV) with vector thrusters, enabling full six-degree-of-freedom (6-DOF) motion. and composite maneuvers. To overcome challenges associated with uncertain model parameters and environmental disturbances, a vision-based tracking control system based on a novel feedforward adaptive model predictive controller (FF-AMPC) is proposed to achieve robust object tracking, which integrates real-time state feedback with adaptive parameter updating. Extensive experiments, including closed-loop tracking and composite motion tests in real-world environment, have been conducted. The obtained results demonstrate that the orientation-adjustable propulsion design significantly improves the underwater maneuverability, while the FF-AMPC controller achieves a 49.1\% hovering improvement compared to conventional controller.
\end{abstract}


%%%%%%%%%%%%%%%%%%%%%%%%%%%%%%%%%%%%%%%%%%%%%%%%%%%%%%%%%%%%%%%%%%%%%%%%%%%%%%%%
\section{Introduction}
\IEEEPARstart{U}{nmanned} underwater vehicles (UUVs) are increasingly used in tasks such as underwater surveying and resource collection due to their high autonomy, strong maneuverability, and wide operational range \cite{Bib:jing2023jmse}. As a robotic system, UUV can perform underwater navigation, obstacle avoidance, and attitude control, which require its capability to estimate its state and adjust its position to reach the target state \cite{Bib:ribas2015auv,Bib:shen2016integrated,Bib:chen2022development}. Notably, three major technical challenges hinder UUV from performing various marine tasks. First, due to the difficulty in accurately modeling time-varying hydrodynamic parameters, UUV is generally unable to obtain accurate estimations of its own state. Second, the environment in which the UUV operates is often characterized by fluctuating water currents, which can interfere with the vehicle's locomotion. Besides, unstructured scenarios require UUV not only to possess full six-degree-of-freedom (6-DOF) control, but also to execute coordinated multi-axis maneuvers such as simultaneous translation and rotation.

Related theories and methods which are appropriate to tackle these challenges have been widely studied in recent years. To overcome the limited omnidirectional mobility and stability of conventional torpedo-shaped UUVs, researchers have proposed various alternative hull configurations. Chen \emph{et al.} \cite{Bib:chen2017computational} proposed a disc-shaped autonomous underwater hull that improves maneuverability during seabed missions. Sean \emph{et al.} \cite{Bib:manley2018aquanaut} developed a morphable-structure UUV that combines flatfish-like swimming capabilities with an expandable frame which, once deployed, functions as a work-class ROV akin to the BlueROV. Soylu \emph{et al.} \cite{Bib:soylu2018sonar} proposed a spherical AUV enabling omnidirectional motion and in-place rotation. Moreover, inspired by piscine locomotion, several researchers have engineered biomimetic underwater robots \cite{Bib:yu2016develop,Bib:tong2022design,Bib:meng2023development} that mimic fish anatomy to achieve enhanced agility or stability. However, most of these biomimetic designs remain small in scale and offer limited thrust, rendering them unsuitable for UUVs intended for real-world operational tasks.

Conventional torpedo-shaped UUVs lack lateral motion, in-place rotation, and hovering capabilities in their standard propulsion configurations. To address these limitations, researchers have proposed several novel propulsion schemes. Sha \emph{et al.} \cite{Bib:sha2025portable} augmented a classical UUV with a multi-thruster system, enabling full 6-DOF motion and in-place maneuvers by distributing thrust across eight propulsors. This multi-propeller method has also become the mainstream approach for both AUVs and ROVs \cite{Bib:richmond2018sunfish,Bib:gelli2018development}. However, a large number of propellers make these vehicles difficult for a single person to handle and drive up their cost. In contrast, vector-thruster design can achieve high maneuverability with fewer propulsors \cite{Bib:maki2018fundamental}, which has attracted significant attention in recent years. For example, Liu \emph{et al.} \cite{Bib:liu2024maneuverable} developed an ROV driven by four vectored thrusters that can perform near-seabed inspection tasks while maintaining stable hovering, although it does not achieve full 6-DOF motion. Jin \emph{et al.} \cite{Bib:jin2015six} introduced a thrust-vectoring ROV (TTURT), capable of complete 6-DOF operation by independently angling each thruster, though some degrees of freedom remain partially coupled.

For the aspect of motion control, although the proportional-integral-derivative (PID) controller features a simple structure, it struggles to cope with the complexities of underwater flow environments and therefore hardly guarantees sufficient stability \cite{bennett1993development}. To address these limitations, researchers have devised a variety of alternative control schemes to enhance UUV stability \cite{Bib:bao2023review}. Among the most widely employed are sliding-mode controller (SMC) \cite{Bib:kong2020iwscr,Bib:cui2016adaptive}, model predictive controller (MPC) \cite{Bib:wang2022adaptive,Bib:zhang2019mpc}, and fuzzy controller \cite{Bib:lin2023depth,Bib:bessa2017design}. Notably, many controllers designed for underwater vehicles emphasize robust state and error tracking. For instance, Wang \emph{et al.} \cite{Bib:wang2024adaptive} integrated an integral term into an SMC framework and adaptively tuned its switching gain, improving pitch angle tracking accuracy. Kong \emph{et al.} \cite{Bib:kong2020extended} incorporated an extended state observer (ESO) within an MPC scheme, which achieved velocity tracking under incomplete state information by modeling motion-tracking errors. Similar adaptive schemes have been reported in several other studies \cite{Bib:sun2024exact,Bib:sun2024adaptive}. Beyond classical model-based techniques, neural-network and reinforcement-learning (RL) approaches have also drawn attention for their ability to learn control policies autonomously \cite{Bib:romero2024actor,Bib:yan2024minimum}. Despite the proliferation of control strategies, there remains a paucity of design studies focused on vehicles employing vector propulsion systems, particularly for hovering, yaw tracking, and target pursuit in vehicles with orientation-adjustable thrusters.

In response to the problems above, a novel underwater vehicle termed orientation-adjustable thruster UUV (OAT-UUV) is designed, followed by corresponding control methods to achieve both high motion performance and stable control capability. The main contributions of this paper are summarized as follows.
\begin{enumerate}
  \item A novel UUV hull is designed by leveraging the structural features of the Myring curve. Comparative simulation experiments have confirmed that the proposed hull structure can support the vehicle's omnidirectional movement and in-place rotation.
  \item An innovative orientation-adjustable propulsion system is introduced, building on vectored-thrust architectures. Both theoretical analysis and experimental results confirm that this system delivers full 6-DOF motion and high maneuverability in real-world environments. To the best of our knowledge, certain composite motion capabilities of the proposed OAT-UUV have never been realized by other vehicles.
  \item A feedforward adaptive model predictive controller (FF-AMPC) is developed based on the proposed propulsion system, which combines predictive feedforward action with adaptive gain updates. Extensive numerical simulations and in-water trials demonstrate the effectiveness of the proposed control scheme, indicating that the controller achieves rapid adjustment and robust stability, offering valuable insights for future ocean exploration and intervention tasks.
\end{enumerate}

The remainder of the paper is organized as follows. Section \uppercase\expandafter{\romannumeral2} provides an overview of the OAT-UUV design and introduces the orientation-adjustable vector propulsion system. At a theoretical level, it is explained how 6-DOF motion can be achieved by adjusting the force and orientation of these thrusters. Section \uppercase\expandafter{\romannumeral3} describes the structure and principles of the FF-AMPC controller. Experimental results and discussions are offered in Section \uppercase\expandafter{\romannumeral4}. Finally, Section \uppercase\expandafter{\romannumeral5} concludes this paper, followed by an outline of future work.

\section{Development of the OAT-UUV}
\label{section2}
\subsection{System Configuration}
\begin{figure}[!t]
    \centering
    \includegraphics[width=0.6\columnwidth]{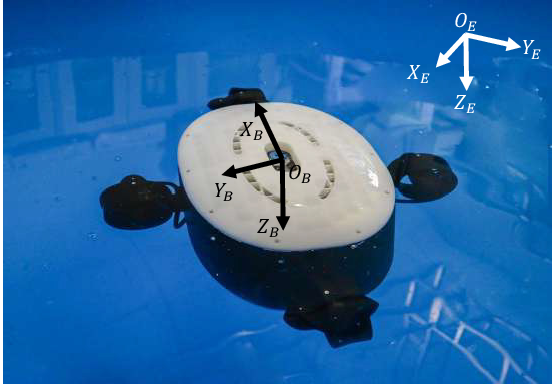}
    \caption{Overview of the OAT-UUV.}
    \label{fig:overview}
\end{figure}

\begin{figure}[!t]
\centering
\includegraphics[width=0.9\columnwidth]{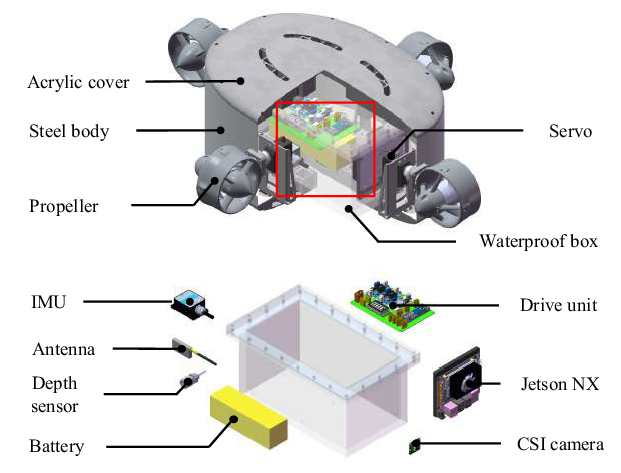}
\caption{Mechatronic configuration of the OAT-UUV.}
\label{fig:mechanical}
\end{figure}

An overview of the proposed OAT-UUV is shown in Fig.~\ref{fig:overview}, with its key mechanical structure shown in Fig.~\ref{fig:mechanical}. Unlike most UUVs, which usually have six or more thrusters, the OAT-UUV is equipped with four orientation-adjustable thrusters, each independently adjusting its direction. By leveraging the orientation-adjustable mechanism of each propeller, the UUV attains propulsion capabilities that extend beyond conventional 6-DOF motion, such as independently actuating all six degrees of freedom and concurrently executing compound maneuvers involving multiple axes. The OAT-UUV measures 60.7 cm$\times$50.2 cm$\times$20.3 cm and weighs 9.2 kg in air, making it easy for single-person handling.

\begin{figure}[!t]
\centering
\includegraphics[width=1.0\columnwidth]{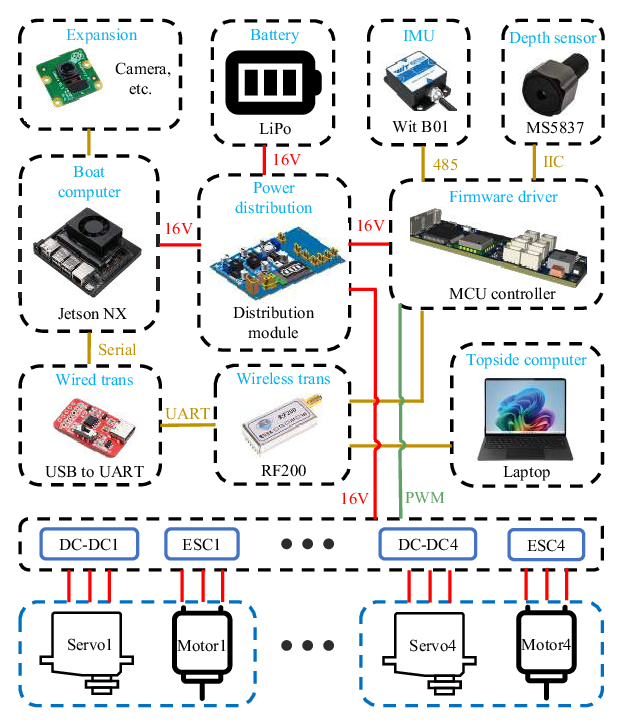}
\caption{Schematic of electrical system of the OAT-UUV.}
\label{fig:schematic}
\end{figure}

The electrical system configuration of OAT-UUV is illustrated in Fig.~\ref{fig:schematic}. Onboard sensors include a camera for horizontal position, an inertial measurement unit (IMU) for orientation, and a pressure sensor for depth estimation. The system is extensible, allowing integration of additional high-end sensors such as sonar or Doppler velocity log (DVL). The OAT-UUV supports both autonomous and remotely operated control modes, namely AUV mode and ROV mode, respectively, thereby broadening its range of applications.

In ROV mode, the firmware driver processes data from the IMU and the depth sensor, which is then transmitted to the topside computer via the antenna. The topside computer performs real-time sensor fusion to estimate the vehicle's state and sends control commands back to the driver unit, which directly drives the four thruster units. The mapping between the PWM signals and the output thrust is established by polynomial curve fitting on the official performance data provided by the thruster manufacturer. AUV mode operates similarly, except that the processed data are sent directly to the onboard computer. The computer fuses these data with its camera inputs to determine the vehicle's state and issues control commands to the firmware driver. The entire system is powered by a battery and features an isolated power module to prevent sensor interference. The technical specifications of the OAT-UUV are detailed in Table~\ref{tab:paraments}.

\begin{table}[!t]
    \centering
    \begin{threeparttable}
    \caption{Technical Parameters of the OAT-UUV}
    \label{tab:paraments}
    \begin{tabular}{c c c c}
        \toprule
        \textbf{Parameters} & \textbf{Value} & \textbf{Parameters} & \textbf{Value} \\  %
        \midrule
        Max speed & 3 $\mathrm{knots}$ & Maneuverability & 0.1 $\mathrm{m}$ \\
        Comm. range & 10 $\mathrm{m}$ & Max depth & 5 $\mathrm{m}$ \\
        Thrust model & T200 & Max duration & 6 $\mathrm{h}$ \\
        \bottomrule  %
    \end{tabular}
    \begin{tablenotes}
    \footnotesize
    \item[1] Maneuverability is measured using the vehicle's horizontal rotation radius.
    \item[2] Communication range is estimated using the vehicle's RF200 firmware parameters.
    \end{tablenotes}
    \end{threeparttable}
\end{table}

\subsection{Hydrodynamic Design of the Myring-Curve Hull}
\begin{figure}[!t]
\centering
\includegraphics[width=0.9\columnwidth]{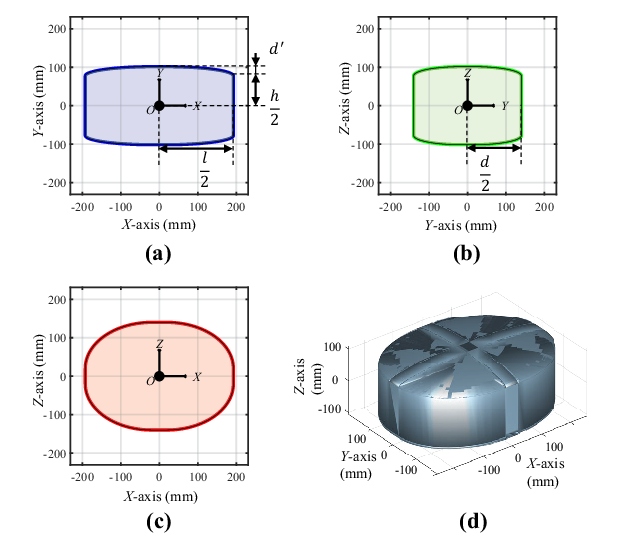}
\caption{Myring curve hull outline of the OAT-UUV. (a) $X-Y$ plane curve. (b) $Y-Z$ plane curve. (c) $X-Z$ plane curve. (d) Envelope surface in 3-D view.}
\label{fig:outline}
\end{figure}

\begin{figure}[!t]
\centering
\includegraphics[width=0.75\columnwidth]{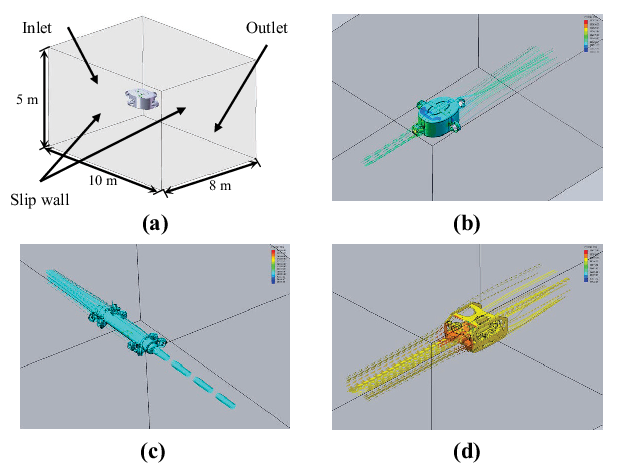}
\caption{Illustration of CFD simulation. (a) Computational domain setup for PMM test simulations. (b) Streamline and static pressure disturbance of OAT-UUV during the PMM test. (c) Streamline and static pressure disturbance of OpenAUV during the PMM test. (d) Streamline and static pressure disturbance of BlueROV during the PMM test.}
\label{fig:pmmtest}
\end{figure}

Since the OAT-UUV is an omnidirectional motion vehicle, it is essential to take the resistance in multiple directions into account during its design. The typical torpedo-like shape used by most UUVs performs poorly in resisting drag in sway and heave directions. Inspired by \cite{Bib:joung2012ijnaoe}, the OAT-UUV's hull is entirely designed with a Myring curve, as shown in Fig.~\ref{fig:outline}. In the coordinate systems shown in Figs.~\ref{fig:outline}(a)--(c), the expressions for the vehicle's respective curves are as follows
\begin{align}
    \label{equ:xycurve} &y=\frac{h}{2}+d'\!\!-\!\!\left[\frac{12d'}{l^2}\!\!-\!\!\frac{2\tan\theta}{l}\right]x^2+\left[\frac{4d'}{l^3}\!\!-\!\!\frac{4\tan\theta}{l^2}\right]x^3,\\
    \label{equ:yzcurve} &z=\frac{h}{2}+d'\!\!-\!\!\left[\frac{12d'}{d^2}\!\!-\!\!\frac{2\tan\theta}{d}\right]y^2+\left[\frac{4d'}{d^3}\!\!-\!\!\frac{4\tan\theta}{d^2}\right]y^3,\\
    \label{equ:xzcurve} &z=\frac{1}{2}d[1-(\frac{2x}{l})^2]^{\frac{1}{2}},
\end{align}
where~$l$, $d$, and $h$ denote the length, width, and height of the OAT-UUV. Note that $d'$ denotes the distance to the highest point of the convex hull. To estimate necessary coefficients, CFD-based Planar Motion Mechanism (PMM) simulations in SOLIDWORKS are conducted. The simulation results are shown in Fig.~\ref{fig:pmmtest}, which indicate that the streamlined hull of the OAT-UUV maintains a laminar flow state at a maximum speed of 3 knots. The CFD simulation utilized a Cartesian-based adaptive mesh, and the flow field was governed by the Reynolds-Averaged Navier-Stokes (RANS) equations coupled with a $k-\epsilon$ turbulence model. For performance comparison, the same CFD analysis is conducted on two representative platforms, OpenAUV and BlueROV, under identical grid density, inlet conditions (Reynolds number~$\approx 10^6$), and turbulence model, with some key indicators shown in Table~\ref{tab:pmmtest}. As can be observed, the OAT-UUV achieves a lower surface pressure compared to OpenAUV and BlueROV in all directions, fully demonstrating the superior hydrodynamic performance of its hull and propulsor layout under complex maneuvering conditions.

\begin{table}[!t]
    %\vspace{-1em}
    \centering
    \caption{Average Static Pressure of Different Vehicles in Three Dimensions}
    \label{tab:pmmtest}
    \begin{tabular}{c c c c}
        \toprule
        \thead{\textbf{Vehicle}} & \thead{\textbf{\textit{X}-\textit{Y} Plane}\\ \textbf{(Pa)}} & \thead{\textbf{\textit{X}-\textit{Z} Plane}\\ \textbf{(Pa)}} & \thead{\textbf{\textit{Y}-\textit{Z} Plane}\\ \textbf{(Pa)}}\\  %
        \midrule
        BlueROV & $\mathrm{1.012}\times \mathrm{10^5}$ & $\mathrm{1.100}\times \mathrm{10^5}$ & $\mathrm{1.011}\times \mathrm{10^5}$\\
        OpenAUV & $\mathrm{1.010}\times \mathrm{10^5}$ & $\mathrm{1.544}\times \mathrm{10^5}$ & $\mathrm{1.013}\times \mathrm{10^5}$\\
        \textbf{OAT-UUV} & $\mathbf{0.999}\times \mathbf{10^5}$ & $\mathbf{1.010}\times \mathbf{10^5}$ & $\mathbf{1.011}\times \mathbf{10^5}$\\
        \bottomrule  %
    \end{tabular}
    %\vspace{-1.5em}
\end{table}

\subsection{Design and Analysis of Orientation Adjustable Propulsion Mechanism}
\label{subsection3}
\begin{figure}[!t]
\centering
\includegraphics[width=0.9\columnwidth]{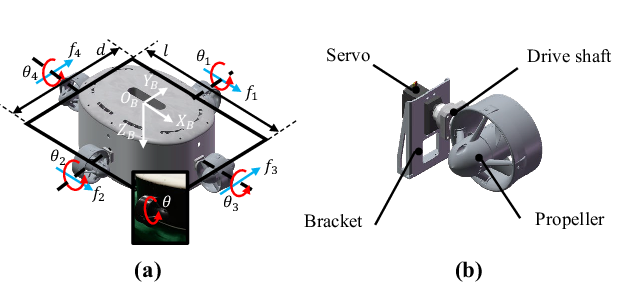}
\caption{Design of orientation adjustable propulsion mechanism. (a) Configuration of thrust vectors. (b) Detailed mechanism of one thruster.}
\label{fig:configuration}
\end{figure}

The OAT-UUV features an orientation-adjustable propulsion mechanism. Four vector thrusters are located at the center of the four edges of the vehicle, with each thruster rotating independently, as illustrated in Fig.~\ref{fig:configuration}(a). The thrust force and moment vectors are determined from the position and orientation of the thrusters, which can be described as
\begin{align}
    \label{vectorgen}
    \tau=\begin{bmatrix}
        f_1\cos\theta_1\!\!+\!\!f_2\cos\theta_2\\
        f_3\cos\theta_3\!\!+\!\!f_4\cos\theta_4\\
        -(f_1\sin\theta_1\!\!+\!\!f_2\sin\theta_2\!\!+\!\!f_3\sin\theta_3\!\!+\!\!f_4\sin\theta_4)\\
        \frac{d}{2}(f_1\sin\theta_1\!\!+\!\!f_2\sin\theta_2)\\
        \frac{d}{2}(f_1\sin\theta_1\!\!+\!\!f_2\sin\theta_2)\\
        \frac{d}{2}(f_1\cos\theta_1\!\!-\!\!f_2\cos\theta_2)\!\!+\!\!\frac{l}{2}(f_3\cos\theta_3\!\!+\!\!f_4\cos\theta_4)
    \end{bmatrix},
\end{align}
where \( f_1 \), \( f_2 \), \( f_3 \), and \( f_4 \) represent the force generated by each thruster, meanwhile \( \theta_1 \), \( \theta_2 \), \( \theta_3 \) and \( \theta_4 \)  represent the tilt angles of each thruster, which are marked in Fig.~\ref{fig:configuration}(a). To achieve independent control over both thrust magnitude and tilt angle for each thruster, an orientation-adjustable mechanism is devised, as shown in Fig.~\ref{fig:configuration}(b). Each propeller mounts directly on its servo shaft, enabling individual rotation driven by a dedicated servo. This mechanism allows each thruster to be actuated independently, thereby meeting the requirements for 6-DOF motion. Premised on identical propulsion hardware dominating the energy budget \cite{Bib:liu2024energy}, a comparative analysis of active thruster count across translational motions indicates that the OAT-UUV reduces average energy consumption by 19.8\% and 33.3\% compared to the BlueROV and OpenAUV, respectively. Meanwhile, analysis based on \cite{Bib:kadiyam2021simulation} indicates that under equivalent propulsion capabilities, the OAT-UUV exhibits a lighter weight, which renders the system especially suitable for high-maneuverability missions.

To quantitatively assess the maneuvering capabilities of the OAT-UUV, a dynamic manipulability analysis (DMA) method is employed\cite{Bib:topini2021dynamic}. Based on classic underwater vehicle modeling methods\cite{Bib:fossen2011handbook}, the kinematic and dynamic equations of the OAT-UUV can be formulated as
\begin{align}
    \label{equ:kinematic} \dot{\eta}&=\mathbf{J}(\eta)\nu,\\
    \label{equ:dynamic} \mathbf{M}\dot{\nu}+\mathbf{C}(\nu)\nu+\mathbf{D}(\nu)\nu+\mathbf{g}(\eta)&=\tau_E+\tau,
\end{align}
where $\eta=[x~y~z~\phi~\theta~\psi]^{\top}\in\mathbb{R}^6$, $\nu=[u~v~w~p~q~r]^{\top}\in\mathbb{R}^6$, and $\tau=[\tau_X~\tau_Y~\tau_Z~\tau_K~\tau_M~\tau_N]^{\top}\in\mathbb{R}^6$ denote the vehicle's pose vector, velocity vector and external thrust vector, respectively. Besides, $\mathbf{M}$, $\mathbf{C}(\nu)$, $\mathbf{D}(\nu)$, and $\mathbf{g}(\eta)$ represent the vehicle's inertia matrix, Coriolis and centripetal terms, hydrodynamic damping matrix and hydrostatic restoring forces. Note that $\tau_E$ accounts for environmental disturbances. Given the configuration of the orientation-adjustable propulsion system shown in Fig.~\ref{fig:configuration}, the vehicle's thrust allocation can be expressed:
\begin{equation}
    \label{allocation}    \tau=\begin{bmatrix}\mathbf{B}_f(\gamma)\\\mathbf{B}_m(\gamma)\end{bmatrix}\begin{bmatrix}f_1&f_2&f_3&f_4\end{bmatrix}^{\top},
\end{equation}
where $B_f(\gamma)$ and $B_m(\gamma)$ represent the vehicle's force and moment allocation matrix, and $\gamma=[\gamma_1~\gamma_2~\gamma_3~\gamma_4]^\top$ denotes the vector of individual thruster orientation angles, respectively. Based on (\ref{equ:kinematic}), (\ref{equ:dynamic}) and (\ref{allocation}), the vehicle's dynamic manipulability Jacobian matrix can be defined as
\begin{equation}
    \label{jacobian}
   \mathbf{J}_{\mathrm{DMA}}(\theta)=\mathbf{M}^{-1}\begin{bmatrix}\mathbf{B}_f(\theta)\\\mathbf{B}_m(\theta)\end{bmatrix}\mathbf{F}_{\max},
\end{equation}
where $F_{\max}=\textrm{diag}(_{\max}f_1,_{\max}f_2,_{\max}f_3,_{\max}f_4)$. To facilitate quantitative evaluation, the manipulability Jacobian matrix is represented as dynamic manipulability ellipsoid (DME). By computing the ellipsoid's volume and eccentricity, the vehicle's overall manipulability and the coupling between its degrees of freedom can be intuitively assessed.

\begin{figure}[!t]
\centering
\includegraphics[width=0.98\columnwidth]{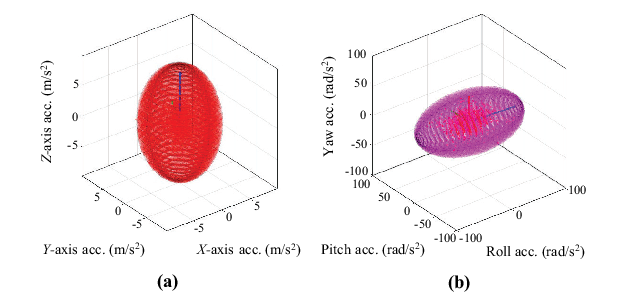}
\caption{Dynamic manipulability ellipsoids of the OAT-UUV. (a) Translational motion DME. (b) Rotational motion DME.}
\label{fig:analysis}
\end{figure}

Utilizing this framework, the OAT-UUV's DME ellipsoids are calculated via MATLAB and illustrated in Fig.~\ref{fig:analysis}. It can be seen that the vehicle attains required acceleration magnitudes in both translational and rotational motions, confirming its strong maneuvering performance. Moreover, the translational and rotational DME eccentricities are 0.83 and 0.67, respectively, which demonstrates the OAT-UUV's excellent isotropy. This implies that the OAT-UUV effectively minimizes coupling among different degrees of freedom while maintaining high maneuverability, thus exhibiting superior agility.

\section{Design of Feedforward Adaptive Controller}
In this section, based on the kinematic and dynamic model of the vehicle, a feedforward adaptive MPC controller (FF-AMPC) is proposed for marine tasks. Combined with the thrust allocation model obtained in Section~\ref{subsection3}, the controller can achieve hovering control and object tracking control of the OAT-UUV. Notably, it exhibits strong robustness against external disturbances.
\subsection{Modeling and Thrust Allocating of OAT-UUV}
As shown in Fig.~\ref{fig:overview}, there is a body-fixed frame \(\mathcal{B} = \{X_B, Y_B, Z_B\}\) attached to the vehicle's center of gravity, and an inertial frame \(\mathcal{I} = \{X_E, Y_E, Z_E\}\) located at a predefined position in the environment. The kinematic and dynamic model of the OAT-UUV is presented in (\ref{equ:kinematic}) and (\ref{equ:dynamic}). Based on the frame definition, the transformation matrix $\mathbf{J}$ can be written as
\begin{equation}
    \label{equ:j}
    \mathbf{J}(\eta)=\begin{bmatrix}\mathbf{J}_1(\phi~\theta~\psi)&\mathbf{0}_{3\times 3}\\\mathbf{0}_{3\times 3}&\mathbf{J}_2(\phi~\theta~\psi)\end{bmatrix},
\end{equation}
where $\mathbf{0}_{3\times 3}$ denotes zero matrix, and the components of $\mathbf{J}$ are given by \cite{Bib:sha2025portable}. Since the vehicle's equilibrium position is at $\theta=0^{\circ}$, the Euler angle singularity condition is assumed to be avoidable throughout the vehicle's maneuvers.

The inertia matrix, incorporating added mass effects, is expressed as $\mathbf{M} = \mathbf{M}_{RB} + \mathbf{M}_{A}$, where $\mathbf{M}_{RB}$ represents the rigid body inertia matrix, and $\mathbf{M}_{A}$ represents the added-mass matrix. The rigid-body inertia matrix is defined by \cite{Bib:fossen2011handbook}, known as
\begin{equation}
    \label{equ:mrb}
    \mathbf{M}_{RB}=\begin{bmatrix}m\mathbf{I}_{3\times 3}&-m\mathbf{S}(r_G)\\m\mathbf{S}(r_G)&\mathbf{I}_B\end{bmatrix},
\end{equation}
where $m$ denotes the vehicle's mass, $\mathbf{I}_{3\times 3}$ is the identity matrix,  $r_G=[x_G~y_G~z_G]^\top\in\mathbb{R}^3$ represents the center of gravity. Note that $\mathbf{I}_B$ denotes the inertia tensor, and $\mathbf{S}(\bullet)$ represents the skew-symmetric operator. In this paper, it is assumed that the OAT-UUV's center of gravity coincides with the center of buoyancy, and the $\tau_E$ can be ignored, thus (\ref{equ:dynamic}) becomes
\begin{equation}
    \label{equ:dynamicsimp}
    \mathbf{M}\dot{\nu}+\mathbf{C}(\nu)\nu+\mathbf{D}(\nu)\nu=\tau.
\end{equation}

Based on (\ref{allocation}), the thrust-allocation scheme for any specific task can be defined. In this paper, two tasks are proposed, including the hovering control and object tracking control. With respect to the hovering control task, $\eta$, $u$, and $\tau$ can be simplified as $\eta=[z~\phi~\theta]$, $\nu=[w~p~q]$, $\tau=[\tau_Z~\tau_K~\tau_M]$. Let $\gamma_i=90^{\circ}$ and $i\in[1,4]$, then the thrust allocation model for this task can be written as
\begin{equation}
    \label{equ:depth_orientation_case}
    \begin{cases}\tau_Z={f_1}+{f_2}+{f_3}+{f_4}\\\tau_K=\frac{d}{2}({f_2}-{f_1})\\\tau_M=\frac{l}{2}({f_4}-{f_3})\end{cases}.
\end{equation}

As for the object tracking control task, $\eta$, $\nu$, and $\tau$ can be simplified as $\eta=[x~y~\psi]$, $\nu=[u~v~r]$, $\tau=[\tau_X~\tau_Y~\tau_N]$. Let \( f_3 = f_4 = 0 \), then the thrust allocation model for this task can be denoted as
\begin{equation}
    \label{equ:trajectory_case}
    \begin{cases}\tau_X=(f_1+f_2)\cos\psi\\
    \tau_Y=(f_1+f_2)\sin\psi\\
    \tau_N=\frac{d}{2}(f_2-f_1)\end{cases}.
\end{equation}

\subsection{Design of FF-AMPC}
\label{sec:ffampc}
The proposed controller represents an enhancement of the baseline MPC, which is executed on the topside computer or the boat computer. Since digital implementation requires a discrete-time formulation, the continuous-time state equations introduced in (\ref{equ:kinematic}) and (\ref{equ:dynamicsimp}) are discretized referring to \cite{Bib:ogata1995prentice}. Meanwhile, it is assumed that the vehicle's state variables are directly observable with negligible errors in subsequent experiments, thus the derivation of component state-space equations can be written as
\begin{align}
    \label{equ:firstmodel} \begin{cases}\eta(k+1)=\mathbf{\Phi}_1\eta(k)+\mathbf{\Gamma}_1\nu(k)\\y_1(k)=\mathbf{I}_{6\times 6}\eta(k)\end{cases},\\
    \label{equ:secondmodel} \begin{cases}\nu(k+1)=\mathbf{\Phi}_2\nu(k)+\mathbf{\Gamma}_2\tau(k)\\y_2(k)=\mathbf{I}_{6\times 6}\nu(k)\end{cases},
\end{align}
where \(\mathbf{I}_{6\times 6}\) is an identity matrix, $\mathbf{\Phi}_1$ and $\mathbf{\Phi}_2$ represent the respective state matrices for position-velocity loop and velocity-thrust loop, and $\mathbf{\Gamma}_1$ and $\mathbf{\Gamma}_2$ denote the input matrices for position-velocity loop and velocity-thrust loop. Note that $\mathbf{\Phi}_1 = \mathbf{I}_{6\times 6}$, $\mathbf{\Gamma}_1 = \mathbf{J}(\eta(k)) \varDelta T$, $\mathbf{\Phi}_2=\varDelta T(\mathbf{I}_{6\times6}-\mathbf{C}(\nu(k))-\mathbf{D}(\nu(k)))/\mathbf{M}$, and $\mathbf{\Gamma}_2=\varDelta T/\mathbf{M}$, where $\varDelta T$ is the sampling time. Meanwhile, $\eta(k)$, $\nu(k)$, and $\tau(k)$ denote the position vector, velocity vector and thrust vector, respectively. Based on (\ref{equ:firstmodel}) and (\ref{equ:secondmodel}), two single-loop MPC are designed for the position-velocity and velocity-thrust components, which are then combined into a complete controller.

Let $x_1(k)=[\varDelta \eta (k)^\top ~y_1(k)^\top]^\top$ and $x_2(k)=[\varDelta \nu(k)^\top ~y_2(k)^\top]^\top$, where $\varDelta \eta(k)=\eta(k)-\eta(k-1)$ and $\varDelta \nu(k)=\nu(k)-\nu(k-1)$. Based on the design theory of classical MPC, (\ref{equ:firstmodel}) and (\ref{equ:secondmodel}) can be transformed into
\begin{equation}
    \label{equ:totalmodel}
    \begin{cases}x_p(k+1)=\mathbf{A}_px_p(k)+\mathbf{B}_p\varDelta u_p(k)\\y_p(k)=\mathbf{C}x_p(k)\end{cases},
\end{equation}
{\setstretch{1.5}where $p\in\{1,2\}$ denotes the symbol of position-velocity loop and velocity-thrust loop. Note that $\mathbf{A}_p=\begin{bmatrix}\mathbf{\Phi}_p&\mathbf{0}_{6\times 6}\\\mathbf{I}_{6\times 6}\mathbf{\Phi}_p&\mathbf{I}_{6\times 6}\end{bmatrix}$,$\mathbf{B}_p=\begin{bmatrix}\mathbf{\Gamma}_p\\\mathbf{I}_{6\times 6}\mathbf{\Gamma}_p\end{bmatrix}$, and $\mathbf{C}=\begin{bmatrix}\mathbf{0}_{6\times 6}&\mathbf{I}_{6\times 6}\end{bmatrix}$, where $\mathbf{0}_{6\times 6}$ denotes zero matrix.}

Above describes the state equations for the entire controller. However, as evident from (\ref{equ:kinematic}) and (\ref{equ:dynamicsimp}), the UUV's state matrices vary with its motion, resulting in state deviations, which cause the conventional MPC to accumulate prediction errors due to delayed parameters. Inspired by \cite{Bib:wang2024adaptive}, a feedforward adaptive component is incorporated, with distinct model adaptation strategies designed for the two control loops.

In the position-velocity component, based on (\ref{equ:kinematic}), it is easy to calculate \(\mathbf{\Phi}_1(k)\) and \(\mathbf{\Gamma}_1(k)\) in real-time based on the data sampled from the system, allowing the model to track the vehicle's motion state in real-time, thus forming a feedforward MPC.

In the velocity-thrust component, the highly nonlinear hydrodynamic forces and moments acting on the UUV are inherently difficult to quantify. Conventional semi-analytical and CFD-based methods are confined to steady-state conditions and typically incur residual modeling errors\cite{Bib:zhao2024research}. Inspired by the work in\cite{Bib:zhu2015adaptive}, an adaptive update session is added to address these issues. Here, the estimated system is designed as
\begin{equation}\label{equ:estimate}
  \hat{x}_2(k+1)=\hat{\mathbf{A}}_2(k)x(k)+\hat{\mathbf{B}}_2(k)\varDelta u_2(k),
\end{equation}
where $\hat{\mathbf{A}}_2(k)$ and $\hat{\mathbf{B}}_2(k)$ are time-varying estimated parameters for uncertain constant matrices $\mathbf{A}_2$ and $\mathbf{B}_2$, as well as $\hat{x}_2(k)$ is the estimated state for the system state $x_2(k)$. Subtracting first equation in (\ref{equ:totalmodel}) and (\ref{equ:estimate}) yields
\begin{equation}\label{equ:error}
  \tilde{x}_2(k+1)=\tilde{\mathbf{A}}_2(k)x_2(k)+\tilde{\mathbf{B}}_2(k)\varDelta u_2(k),
\end{equation}
where $\tilde{x}_2(k)=x_2(k)-\hat{x}_2(k)$, $\tilde{\mathbf{A}}_2(k)=\mathbf{A}_2(k)-\hat{\mathbf{A}}_2(k)$, and $\tilde{\mathbf{B}}_2(k)=\mathbf{B}_2(k)-\hat{\mathbf{B}}_2(k)$. Let $\hat{\mathbf{\Theta}}_2(k) = \begin{bmatrix}\hat{\mathbf{A}}_2(k) & \hat{\mathbf{B}}_2(k)\end{bmatrix}$, and $X(k) = \begin{bmatrix}x_2(k)^{\top} & \varDelta u_2(k)^\top\end{bmatrix}^\top$, then the adaptive update law is written as
\begin{equation}
    \label{equ:adaptive}
    \hat{\mathbf{\Theta}}_2(k+1)=\hat{\mathbf{\Theta}}_2(k)+\lambda\tilde{x}_2(k+1)X(k)^\top ,
\end{equation}
where $\lambda \ge 0$ is the updating rate to be assigned. In a control cycle of the velocity-thrust loop, the input speed feedback value is first used to compute $X(k)$ and $\tilde{x}(k)$. These values are then used to calculate the updated $\hat{\mathbf{A}}_2(k)$ and $\hat{\mathbf{B}}_2(k)$ via (\ref{equ:adaptive}), which are subsequently fed into the velocity-thrust MPC to compute the control output.

As for the cost function $\mathbf{J}_p$, since the control task is to minimize the track errors, a weighted least squares tracking error is added along with the control cost. In the FF-AMPC, $N_p$ is set as the prediction horizon, and $N_c$ as the control horizon, meanwhile $\bar{r_1}$ and $\bar{r_2}$ are the weight parameters of error cost and energy cost which need to be assigned manually, and $r_p$ denotes the reference signal in each loop. Therefore, the discrete-time cost over the prediction horizon is written as
\begin{equation}
    \label{equ:cost}
    \begin{aligned}
    \mathbf{J}_p(k)&=(Y_p(k)-{R_p}(k))^\top\bar{\mathbf{R}_1}(Y_p(k)-{R_p}(k))\\ &+\varDelta U_p(k)^\top\bar{\mathbf{R}_2}\varDelta U_p(k),
    \end{aligned}
\end{equation}
where $\bar{\mathbf{R}_1}=\mathrm{diag}(\bar{r_1})\in\mathbb{R}^{N_c\times N_c}$, $\bar{\mathbf{R}_2}=\mathrm{diag}(\bar{r_2})\in\mathbb{R}^{N_c\times N_c}$, and $R_p(k)=[{r_p(k+1)^\top},\dots,r_p(k+N_p)^\top]^\top$. Note that $Y_p(k)=[y_p(k+1|k)^\top,\dots,y_p(k+N_p|k)^\top]^\top$ and $\varDelta U_p(k)=[\varDelta u_p(k), \dots,\varDelta u_p(k+N_c-1|k)^\top]^\top$, meanwhile $p\in\{1,2\}$ denotes the position-velocity loop and velocity-thrust loop. To guarantee the stability of the closed-loop estimated system, a terminal constraint is written as
\begin{equation}
    \label{equ:terminal}
    y_p(k+N_c|k)=r_p(k+N_c|k).
\end{equation}

Meanwhile, to ensure the feasibility of the adaptive update law, a constraint condition is also added, written as
\begin{equation}
    \label{equ:adaptive_constraint}
    X(k)^\top X(k)<\frac{2-\alpha}{\lambda},
\end{equation}
where \(\alpha \in [0, 2]\). The stability of the proposed FF-AMPC scheme is theoretically guaranteed via Lyapunov analysis \cite{Bib:grune2016nonlinear}, which demonstrates the asymptotic convergence of tracking errors and the boundedness of adaptive parameters. In the implementation, $\lambda$ is first selected to guarantee convergence, followed by the adjustment of $\alpha$ to suppress overshoot. After that, joint fine-tuning is performed experimentally to optimize tracking performance.

In brief, the FF-AMPC controller can be summarized in the position-velocity loop and velocity-thrust loop with adaptive laws, written as
\begin{align}
    \label{equ:pvmpc}\begin{cases}\mathbf{\Phi}_1(k)=\mathbf{I}_{6\times6},\mathbf{\Gamma}_1(k)=\mathbf{J}(\eta(k))\Delta T\\x_1(k+1)=\mathbf{A}_1(k)x_1(k)+\mathbf{B}_1(k)\varDelta u_1(k)\\\varDelta U_1(k)=\arg\min\limits_{u_1}\mathbf{J}_1(k)\\\varDelta u_1(k)=[\mathbf{I}_{6\times 6},\mathbf{0}_{6\times 6},\dots,\mathbf{0}_{6\times 6}]\varDelta U_1(k)\\y_1(k+N_c|k)=r_1(k+N_c|k)\end{cases},\\
    \label{equ:vtmpc}\begin{cases}\hat{\mathbf{\Theta}}_2(k)=\hat{\mathbf{\Theta}}_2(k-1)+\lambda\tilde{x}_2(k)X(k-1)^\top\\\hat{x}_2(k+1)=\hat{\mathbf{A}}_2(k)x_2(k)+\hat{\mathbf{B}_2}(k)\varDelta u_2(k)\\\varDelta U_2(k)=\arg\min\limits_{u_2}\mathbf{J}_2(k)\\\varDelta u_2(k)=[\mathbf{I}_{6\times 6},\mathbf{0}_{6\times 6},\dots,\mathbf{0}_{6\times 6}]\varDelta U_2(k)\\X(k)^\top X(k)<\frac{2-\alpha}{\lambda}\\y_2(k+N_c|k)=r_2(k+N_c|k)\end{cases}.
\end{align}

The total controller can be divided into two loops, expressed as (\ref{equ:pvmpc}) and (\ref{equ:vtmpc}). The first row of each loop is the adaptive law while the second row denotes the state equation after updating the parameter matrices, as well as the third and the fourth rows denote the calculation of control sequences. The rest of each loop is the constraints, making sure the adaptive update law be hold, and the control sequences bounded. Instead of relying on fixed hydrodynamic parameters which are difficult to identify accurately, the integrated adaptive law continually updates the system matrices to implicitly compensate for actual hydrodynamic forces and unmodeled dynamics by minimizing real-time prediction errors.

\subsection{Vision-Based Tracking}
\begin{figure}[!t]
    \centering
    \includegraphics[width=0.98\columnwidth]{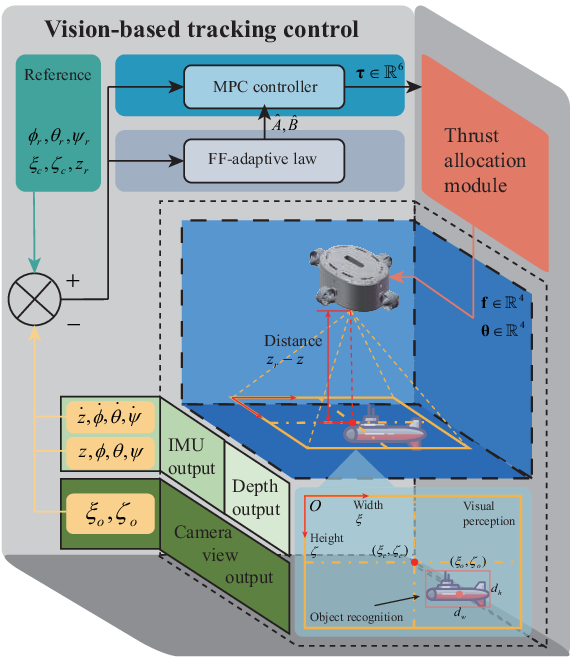}
    \caption{Vision-based tracking control diagram and FF-AMPC framework for the OAT-UUV.}
    \label{fig:framework}
    %\vspace{-1em}
\end{figure}

The vision-based tracking module is built upon the FF-AMPC control architecture detailed in Section~\ref{sec:ffampc} and the thrust-allocation scheme of Section~\ref{subsection3}, with its framework illustrated in Fig.~\ref{fig:framework}. An overhead camera is used to capture video frames. To minimize measurement errors caused by lens distortion, its intrinsic matrix is pre-calibrated and utilized. Each frame is converted to the hue-saturation-value (HSV) color space and filtered via OpenCV routines to segment the target object, where the center of the frame $O_c=(\xi_c,\zeta_c)$ and the pixel coordinates $(\xi,\zeta)$ are defined. The outputs of the object detection module include the object's bounding box center, width, and height, which are denoted as $O_o=(\xi_o,\zeta_o)$,~$d_w$, and $d_h$. Here, the normalized errors along $X$ axis and $Y$ axis are given by
\begin{equation}\label{normerrors}
  \tilde{e}_w=\frac{\xi_o-\xi_c}{d_w},~\tilde{e}_h=\frac{\zeta_o-\zeta_c}{d_h}.
\end{equation}

This visual feedback is used to control surge and heading, and the sensor feedback is used to control depth, roll angle and pitch angle. Here, the feedback values from the sensors are converted into the vehicle's pose and velocity state feedback values, denoted as $\eta_f=[\xi_f~\zeta_f~z_f~\phi_f~\theta_f~\psi_f]$ and $u_f=[\dot{\xi_f}~\dot{\zeta_f}~\dot{z_f}~\dot{\phi_f}~\dot{\theta_f}~\dot{\psi_f}]$, which are then fed into the controller along with the pose target written as \(\eta_r=[\xi_r~\zeta_r~z_r~\phi_r~\theta_r~\psi_r]\). In a complete control loop, the feedback values are first used to update the estimated states, which are then combined with the reference values and input into the state-updated controller to compute the control output. Note that the output computed from the first sub-module is used as velocity target $u_r=[\dot{\xi_r}~\dot{\zeta_r}~\dot{z_r}~\dot{\phi_r}~\dot{\theta_r}~\dot{\psi_r}]$ and fed into the second sub-module. The final computed total thrust and torque \(\mathbf{\tau}\) are then sent to the thrust allocation module, which calculates the force vector $\mathbf{f}$ and orientation vector $\mathbf{\theta}$ for UUV.

\section{Experimental Results}
To validate the motion capabilities and controller advantages of the OAT-UUV, a series of open-loop motion and closed-loop control experiments are performed with the OAT-UUV prototype. The experimental setup involves a $\phi~3~\mathrm{m}$ pool with $0.7~\mathrm{m}$ height. In these experiments, the maneuverability of the vehicle, the robustness of the controller, and its tracking performance are comprehensively evaluated. It should be noted that since black provides superior recognition performance in the global camera, the UUV's paint scheme was changed during the experiment to achieve better experimental results.

\subsection{Open-Loop Maneuverability}
\begin{figure}[!t]
    \centering
    \includegraphics[width=0.98\columnwidth]{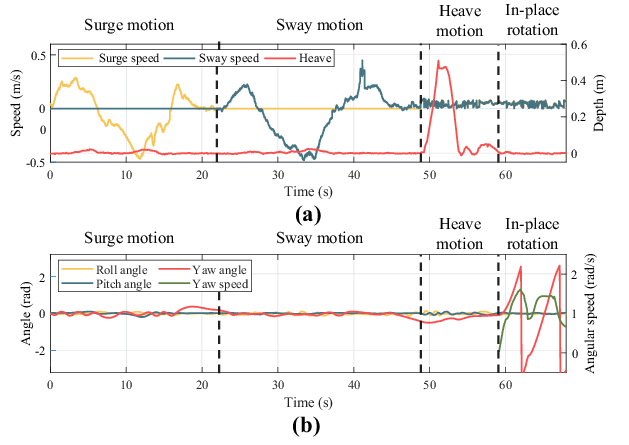}
    \caption{Tendency curves of open-loop motion experiment in real-world environment. (a) Linear speed and depth. (b) Orientation angle and angular speed.}
    \label{fig:openloop_curve}
\end{figure}

\begin{figure}[!t]
    \centering
    \includegraphics[width=0.9\columnwidth]{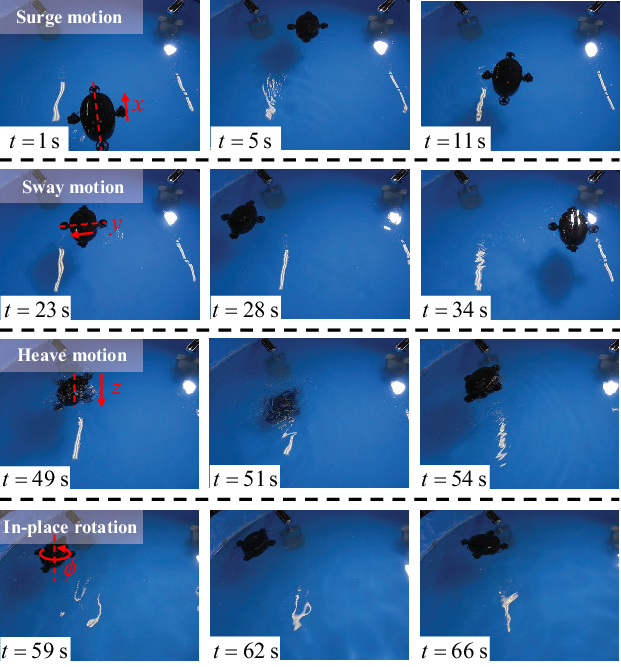}
    \caption{Video snapshots of open-loop motion process.}
    \label{fig:openloop_video_snaps}
\end{figure}

First, the performance in linear and rotational maneuvers of OAT-UUV is examined under open-loop motion. To quantify its maneuverability, steady-state velocity and minimum turning radius are measured, as plotted in Figs.~\ref{fig:openloop_curve}~(a) and (b). The data reveal a persistent yaw offset during both surge and sway operations, indicating that motion errors are unavoidable in open-loop mode. A heading calibration logic is incorporated into it, enabling it to maintain good heading accuracy. While maintaining this heading deviation, the vehicle reaches a maximum horizontal speed of 0.4~m/s, with its heave velocity slightly exceeding this value. In the rotation motion, the vehicle achieves a rotation speed of 1.5~rad/s and the minimum turning radius is 0.05~m, which is substantially smaller than the vehicle's overall dimensions. Video snapshots of each motion are provided in Fig.~\ref{fig:openloop_video_snaps}.

\begin{figure}[!t]
    \centering
    \includegraphics[width=0.98\columnwidth]{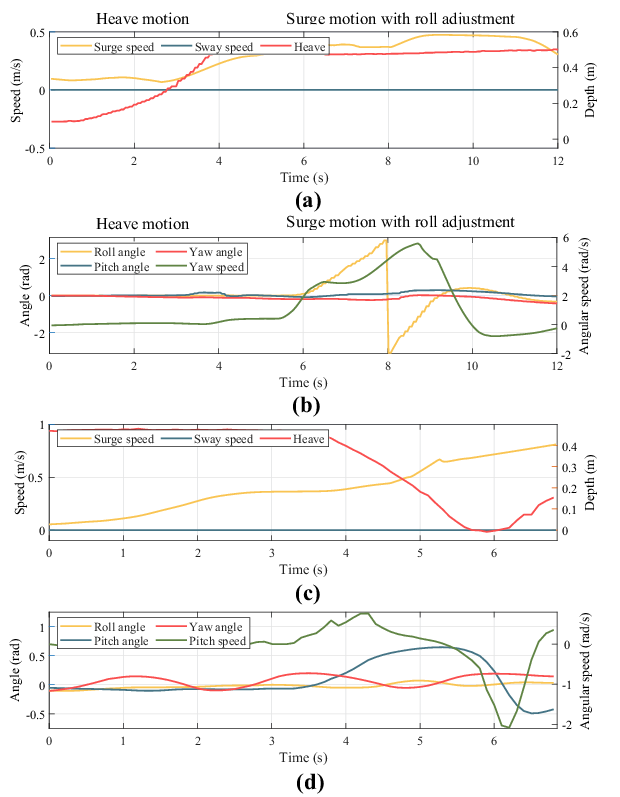}
    \caption{Tendency curves of combined motion experiment in real-world environment. (a) Linear speed and depth in heave-surge-roll combined motion. (b) Orientation angle and angular speed in heave-surge-roll combined motion. (c) Linear speed and depth in surge-pitch combined motion. (d) Orientation angle in surge-pitch combined motion.}
    \label{fig:combloop_curve}
\end{figure}

\begin{figure}[!t]
    \centering
    \includegraphics[width=0.9\columnwidth]{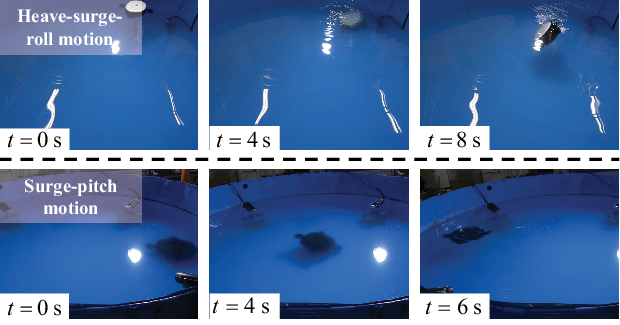}
    \caption{Video snapshots of combined motion process.}
    \label{fig:combloop_video_snaps}
\end{figure}

In addition, the multi-DOF maneuverability of the OAT-UUV is evaluated by two open-loop combined-motion tests. The first one is a heave-surge-roll combined motion experiment. The vehicle submerges first and then advances forward while continuously adjusting its roll angle, emulating the posture changes required to negotiate a narrow passage in a real-world environment. The second one is a surge-pitch combined motion experiment. The vehicle moves forward while adjusting its pitch angle, which is a common maneuver in underwater exploration. Motion parameters of the maneuver are provided in Fig.~\ref{fig:combloop_curve}, with snapshots presented in Fig.~\ref{fig:combloop_video_snaps}. Experimental results show that the OAT-UUV can perform continuous posture adjustments while maintaining a stable speed. No sharp angle change is observed during translations, indicating that it can keep stable during combined motion, thereby validating the platform's ability to execute multi-DOF maneuvering tasks in complex underwater environments.

\subsection{Hovering}
\begin{figure}[!t]
    \centering
    \includegraphics[width=0.9\columnwidth]{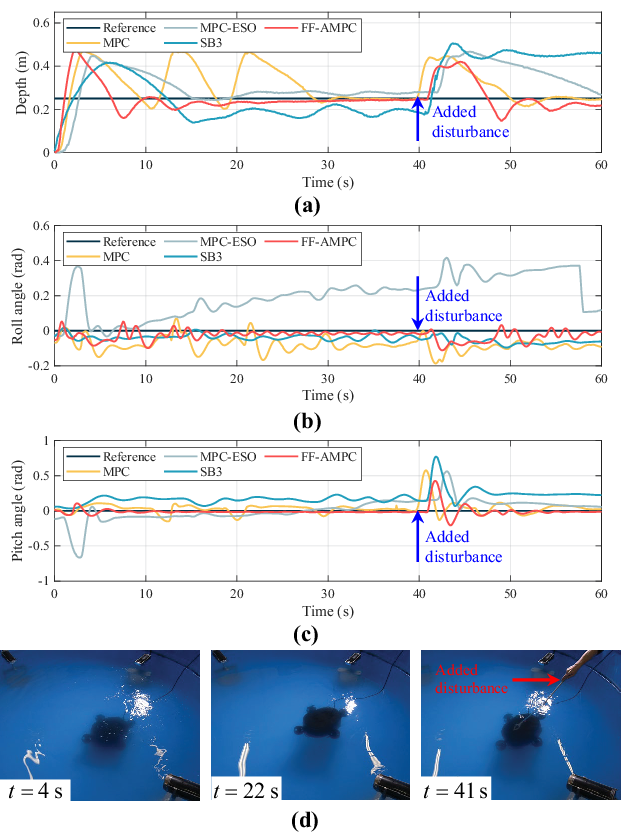}
    \caption{Experiments on hovering control in simulation and real environment. (a) Tendency curve of depth. (b) Tendency curve of roll angle. (c) Tendency curve of pitch angle. (d) Video snapshots of the hovering processes using FF-AMPC.}
    \label{fig:hovering}
\end{figure}

This subsection reports and analyzes the performance of the proposed FF-AMPC on hovering motion. The evaluation consists of two parts. First, the vehicle begins with its initial state and tracks a reference trajectory, in which four controllers are compared including MPC \cite{Bib:wang2009springer}, MPC-ESO \cite{Bib:kong2020extended}, Stable-baseline3 (SB3) \cite{Bib:raffin2021stable} and FF-AMPC. At the 40th second mark of the experiment, an external instantaneous disturbance of equal magnitude is applied at the same position on the OAT-UUV to evaluate the disturbance rejection capabilities of the four controllers. The initial trajectory for all tests is defined as $[z_0~\phi_0~\theta_0]=[0~\mathrm{m}~0~\mathrm{rad}~0~\mathrm{rad}]$, with the reference trajectory set as $[z_r~\phi_r~\theta_r]=[0.25~\mathrm{m}~0~\mathrm{rad}~0~\mathrm{rad}]$. Tracking accuracy is evaluated using the root mean square error (RMSE), while disturbance rejection capability is assessed via the integral absolute error (IAE) following disturbance application. Each experiment lasts 60~s with control loop operating at 10~Hz. Note that Figs.~\ref{fig:hovering}(a)--(c) display the tracking trajectories, while Fig.~\ref{fig:hovering}(d) shows representative snapshots from the FF-AMPC physical experiment, including the process of applying the disturbance. Concurrently, the numerical performance metrics obtained from each experiment are summarized in Tables~\ref{tab:hovering_rmse} and~\ref{tab:hovering_iae}. It is evident that FF-AMPC achieves the minimum total RMSE and minimum IAE under multi-degree-of-freedom conditions, demonstrating its capability to effectively resist external disturbances and attenuate errors to minimal steady-state values in the presence of interference. Notably, the SB3 algorithm yielded a higher RMSE and failed in disturbance rejection tests. This underperformance is attributed to the significant sim-to-real gap caused by the lack of a high-fidelity hydrodynamic physics engine in the training environment. This confirms that the proposed model-based approach is currently more robust for practical deployment.

\begin{table}[htbp]
\centering
\caption{RMSE for Different Controllers in Hovering Task}
\label{tab:hovering_rmse}
\begin{tabular}{c c c c}
\toprule
\multirow{2}{*}{\textbf{Method}} &\multicolumn{3}{c}{\textbf{RMSE}}\\
\cmidrule(lr){2-4}
& Depth (m) & Roll (rad) & Pitch (rad)\\
\midrule
MPC~\cite{Bib:wang2009springer} & 0.1104 & 0.0882 & 0.0934\\
MPC-ESO~\cite{Bib:kong2020extended} & 0.1087 & 0.2303 & 0.1536 \\
SB3~\cite{Bib:raffin2021stable} & 0.1348 & 0.0483 & 0.2142 \\
\textbf{FF-AMPC}& \textbf{0.0704} & \textbf{0.0391} & \textbf{0.0617} \\
\bottomrule
\end{tabular}
\end{table}

\begin{table}[htbp]
\centering
\caption{IAE for Different Controllers After Applying Disturbance}
\label{tab:hovering_iae}
\begin{tabular}{c c c c}
\toprule
\multirow{3}{*}{\textbf{Method}} &\multicolumn{3}{c}{\textbf{IAE}}\\
\cmidrule(lr){2-4}
& Depth & Roll & Pitch\\
& ($\mathrm{m} \cdot \mathrm{s}$) & ($\mathrm{rad} \cdot \mathrm{s}$) & ($\mathrm{rad} \cdot \mathrm{s}$)\\
\midrule
MPC~\cite{Bib:wang2009springer} & --1.1395 & 1.8564 & --1.2110\\
MPC-ESO~\cite{Bib:kong2020extended} & --2.4171 & --5.6189 & --2.3762 \\
SB3~\cite{Bib:raffin2021stable} & --3.7521 & 1.1722 & --5.2170 \\
\textbf{FF-AMPC}& \textbf{--0.3283} & \textbf{0.7420} & \textbf{--0.0752} \\
\bottomrule
\end{tabular}
\end{table}

\subsection{Object Tracking}
\begin{figure}[!t]
    \centering
    \includegraphics[width=0.9\columnwidth]{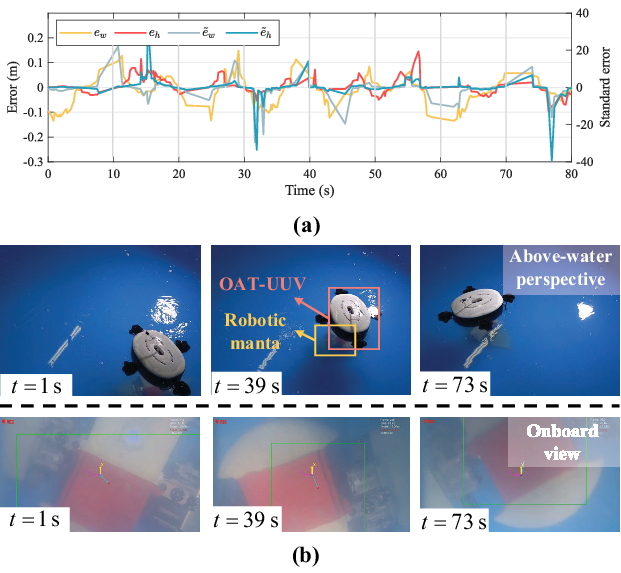}
    \caption{Experiments on object tracking. (a) Tendency curve of $e_w$, $e_h$, $\tilde{e}_w$ and $\tilde{e}_h$. (b) Video snapshots of the object tracking process.}
    \label{fig:objecttracking}
\end{figure}

This subsection reports and analyzes the vision-based autonomous locomotion capability of the OAT-UUV in a real-world environment. The camera fixed on the bottom of OAT-UUV is used to detect and continuously track a moving robotic manta\cite{Bib:meng2020toward}, with a red patch affixed to the manta's dorsal surface serving as the visual marker. For quantitative evaluation, the standard tracking error $\tilde{e}_w$ and $\tilde{e}_h$, and absolute tracking error $e_w=\tilde{e}_w\times d_w$ and $e_h=\tilde{e}_h\times d_h$ are measured along the horizontal and vertical axes within the camera's field of view (FOV), as plotted in Fig.~\ref{fig:objecttracking}(a). It can be noted that during phases of steady target motion, the tracking error remains low, and transient peaks appear when the target executes turns, but the system recovers to  steady-state within a short time. Fig.~\ref{fig:objecttracking}(b) presents experimental snapshots from multiple views to visually illustrate the tracking process and dynamic response of the OAT-UUV. To handle target loss events, a safety-prioritized re-acquisition logic is implemented. Upon losing visual contact, the UUV moves toward the target's last observed coordinates. If re-acquisition fails within a 3-second threshold, the controller aborts the search and engages an idle mode to prevent collision and drift, awaiting operator intervention.

\subsection{Discussion}
The effectiveness of the OAT-UUV's design and control is confirmed across several key performance areas. First, the various motions and combined ones, as shown in Figs.~\ref{fig:openloop_video_snaps} and \ref{fig:combloop_video_snaps}, indicate the vehicle's ability to perform superior maneuvers while offering excellent portability and reducing coupling between degrees of freedom. This unique capability, which is a direct result of the orientation-adjustable thruster design, confirms the platform's high maneuverability for navigating complex spaces. Besides, the hovering trials quantitatively demonstrate the superiority of the FF-AMPC controller. As shown in Fig.~\ref{fig:hovering} and the errors in Tables~\ref{tab:hovering_rmse} and \ref{tab:hovering_iae}, FF-AMPC achieves significantly lower steady-state errors and faster settling times as opposed to other MPC-based and RL-based controllers. This highlights its capacity to adapt to complicated environment through real-time model updates, a critical feature for robust operation. Finally, the vision-based object tracking experiment, as shown in Fig.~\ref{fig:objecttracking}, confirms the successful integration of the entire system. The ability to autonomously follow a dynamic target in real time not only validates the robustness of the complete perception-to-control loop in a practical application but also shows the platform's potential for interactive missions.

Despite these promising results, the experiments also highlight current limitations and areas for improvement. For instance, while the vision-based tracking system is effective, it can be observed from Fig.~\ref{fig:objecttracking} that the camera has a small FOV, implying that the target may leave the field of view during the tracking process. Furthermore, although the closed-loop controller effectively stabilizes the vehicle, residual errors persist due to hardware imperfections in the handmade prototype, where assembly tolerances introduce parasitic disturbance torques. Mitigating these mechanical uncertainties through precision manufacturing and data-driven compensation techniques could improve overall efficiency and performance. Moreover, it faces the accumulation of heading drift during extended maneuvers due to the reliance on gyroscope integration, which will be addressed in future work by fusing data from other absolute orientation references. Most significantly, all experiments were performed in the controlled environment of a laboratory pool. The vehicle's performance against more dynamic and unpredictable disturbances, such as strong water currents and poor visibility found in real-world aquatic environments, has yet to be evaluated. The robustness of the current control and vision systems in such challenging conditions remains a critical area for future investigation.

\section{Conclusion and Future Work}
This paper proposes and implements the OAT-UUV with enhanced maneuverability and adaptability. First, a new orientation adjustable propulsion system is developed, which enables the vehicle to perform 6-DOF motions as well as combined motions across multiple degrees of freedom. Second, in response to the challenges posed by the difficulty of estimating the actual state of the vehicle and the complexity and variability of the environment, FF-AMPC is proposed to enhance the control capability of the OAT-UUV. Eventually, extensive experiments are carried out to verify the vehicle's superior motion capabilities and tracking performance.

Future work will explore learning-based control to enhance robustness for complex inspections under extreme disturbances. Integrating sensors like DVL and sonar will further enhance multimodal perception, enabling the OAT-UUV to execute precise tasks in confined, real-world aquatic environments, thereby maximizing its practical value.

\vspace{-1cm}%
\begin{IEEEbiography}[{\includegraphics[width=1in,height=1.25in,clip,keepaspectratio]{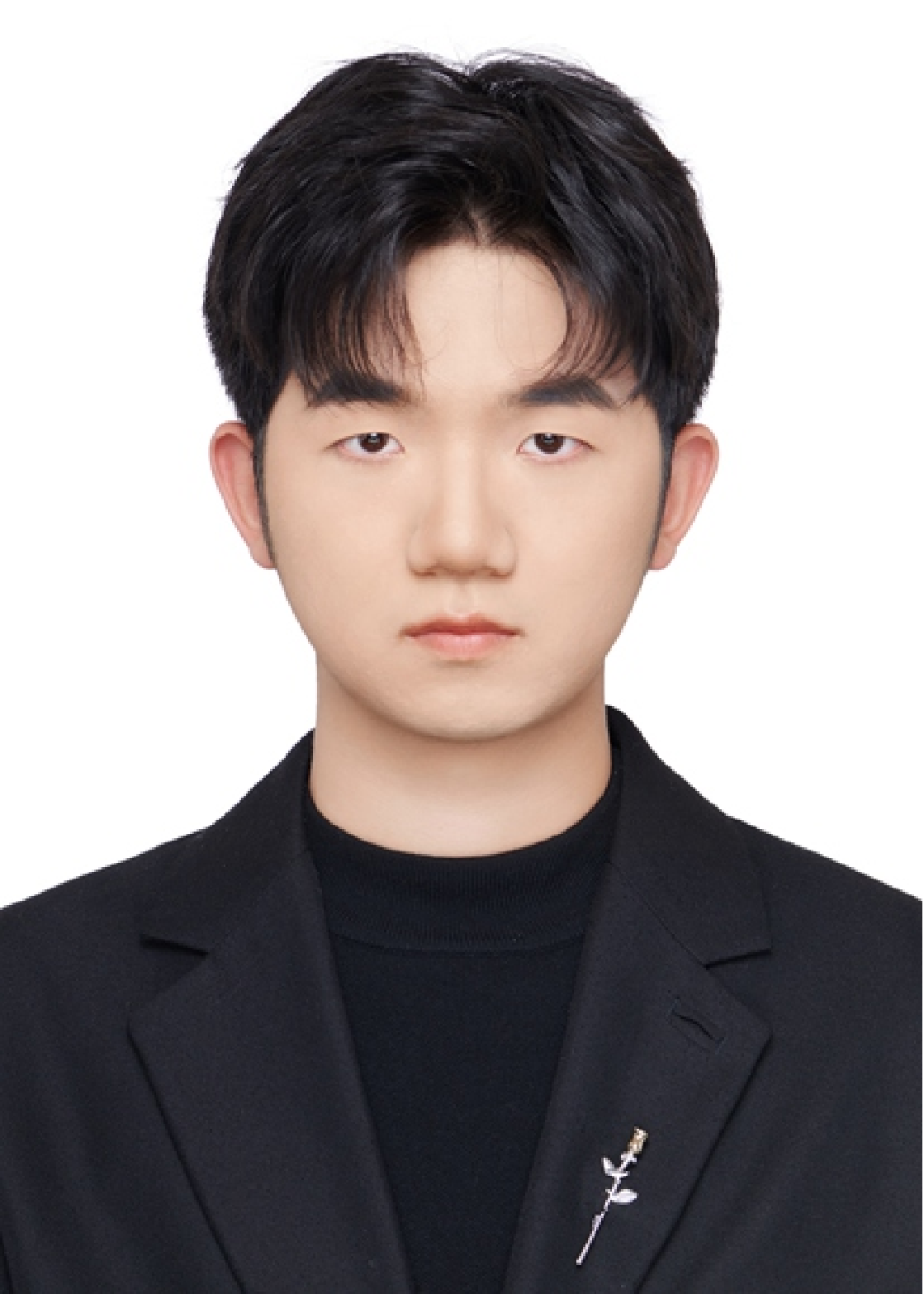}}]{Yifei Wang} received the B.E. degree in science in intelligent manufacturing engineering from the School of Mechanical Engineering, Beijing Institute of Technology, Beijing, China, in 2024. He is currently pursuing the Ph.D. degree in general mechanics and foundation of mechanics with the School of Advanced Manufacturing and Robotics, Peking University, Beijing.

His current research interests include underwater robotics and synaesthetic perception.
\end{IEEEbiography}

\vspace{-1cm}
\begin{IEEEbiography}[{\includegraphics[width=1in,height=1.25in,clip,keepaspectratio]{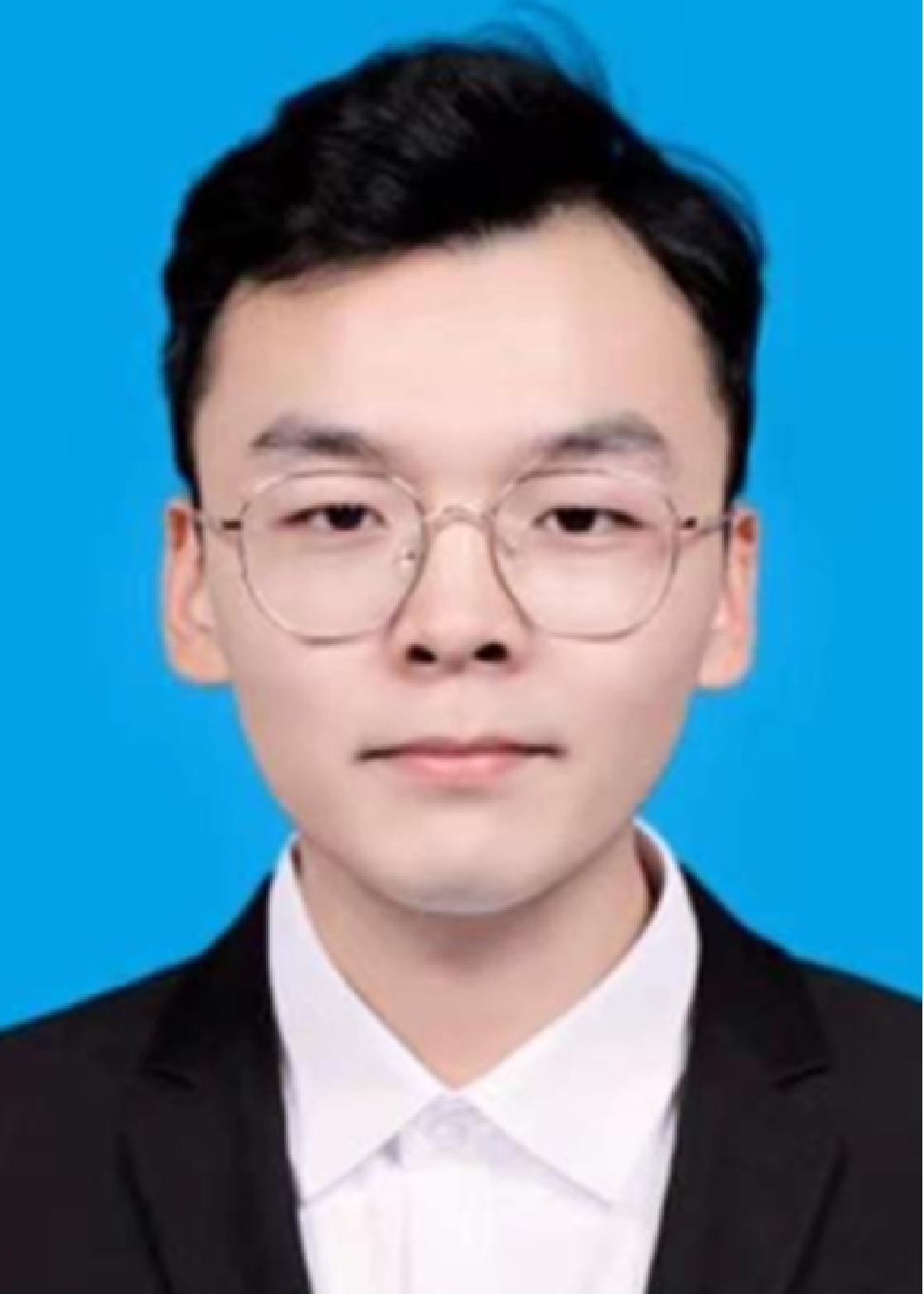}}]{Shihan Kong} (Member, IEEE) received the B.E. degree in automation from the School of Control Science and Engineering, Shandong University, Jinan, China, in 2016, and the Ph.D. degree in control theory and control engineering from the Institute of Automation, Chinese Academy of Sciences, Beijing, China, in 2021.

He is currently a Research Assistant Professor with the School of Advanced Manufacturing and Robotics, Peking University, Beijing. His current research interests include underwater robotics and robot vision.
\end{IEEEbiography}

\vspace{-1cm}
\begin{IEEEbiography}[{\includegraphics[width=1in,height=1.25in,clip,keepaspectratio]{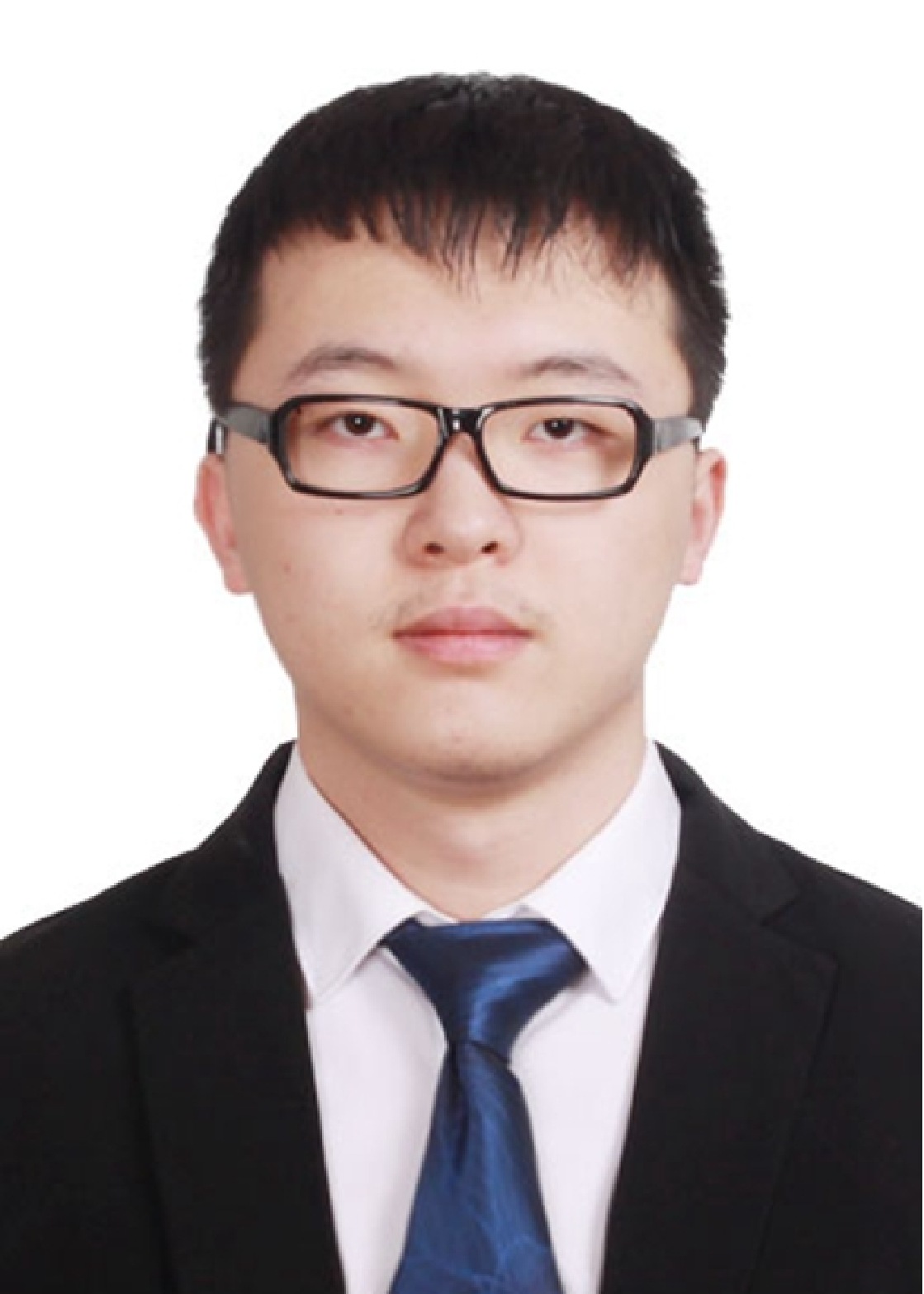}}]
{Zhanhua Xin}
received the B.E. degree in mechatronic engineering from the School of Xuteli, Beijing Institute of Technology, Beijing, China, in 2022. He is currently pursuing the Ph.D. degree in general mechanics and foundation of mechanics with the School of Advanced Manufacturing and Robotics, Peking University, Beijing.

His research interests include computer vision and bioinspired underwater robots.
\end{IEEEbiography}

\vspace{-1cm}%
\begin{IEEEbiography}[{\includegraphics[width=0.95in,height=1.25in,clip,keepaspectratio]{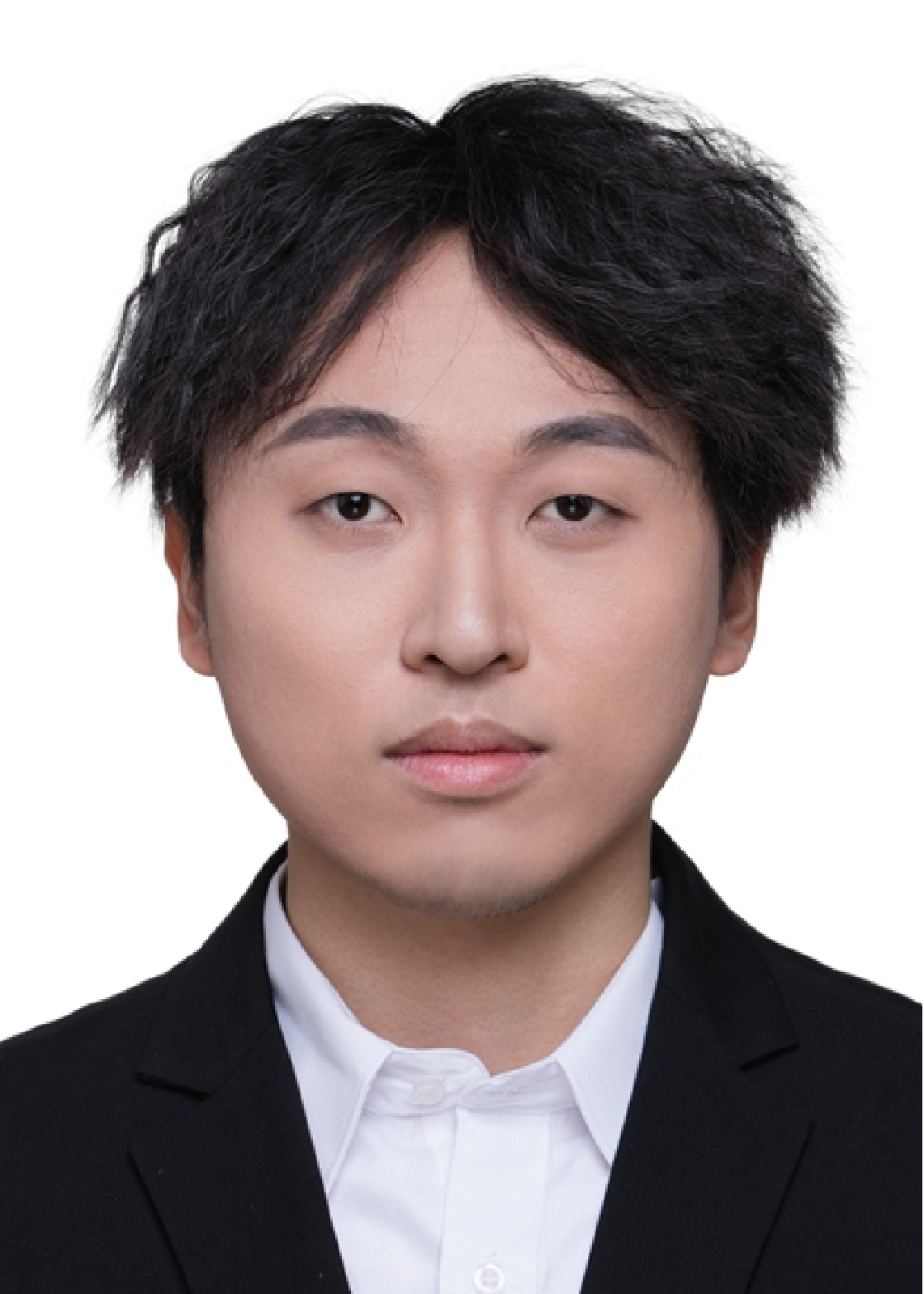}}]{Kaiwei Zhu} received the B.S. degree in ocean engineering and technology from the Ocean College, Zhejiang University, Hangzhou, China, in 2022. He is currently working toward the M.S. degree in mechanical systems and control with the School of Mechanics and Engineering Science, Peking University, Beijing, China.

His current research interests include intelligent control, motion planning, and unmanned surface vehicle.
\end{IEEEbiography}

%\vspace{-1cm}%
\begin{IEEEbiography}[{\includegraphics[width=1in,height=1.25in,clip,keepaspectratio]{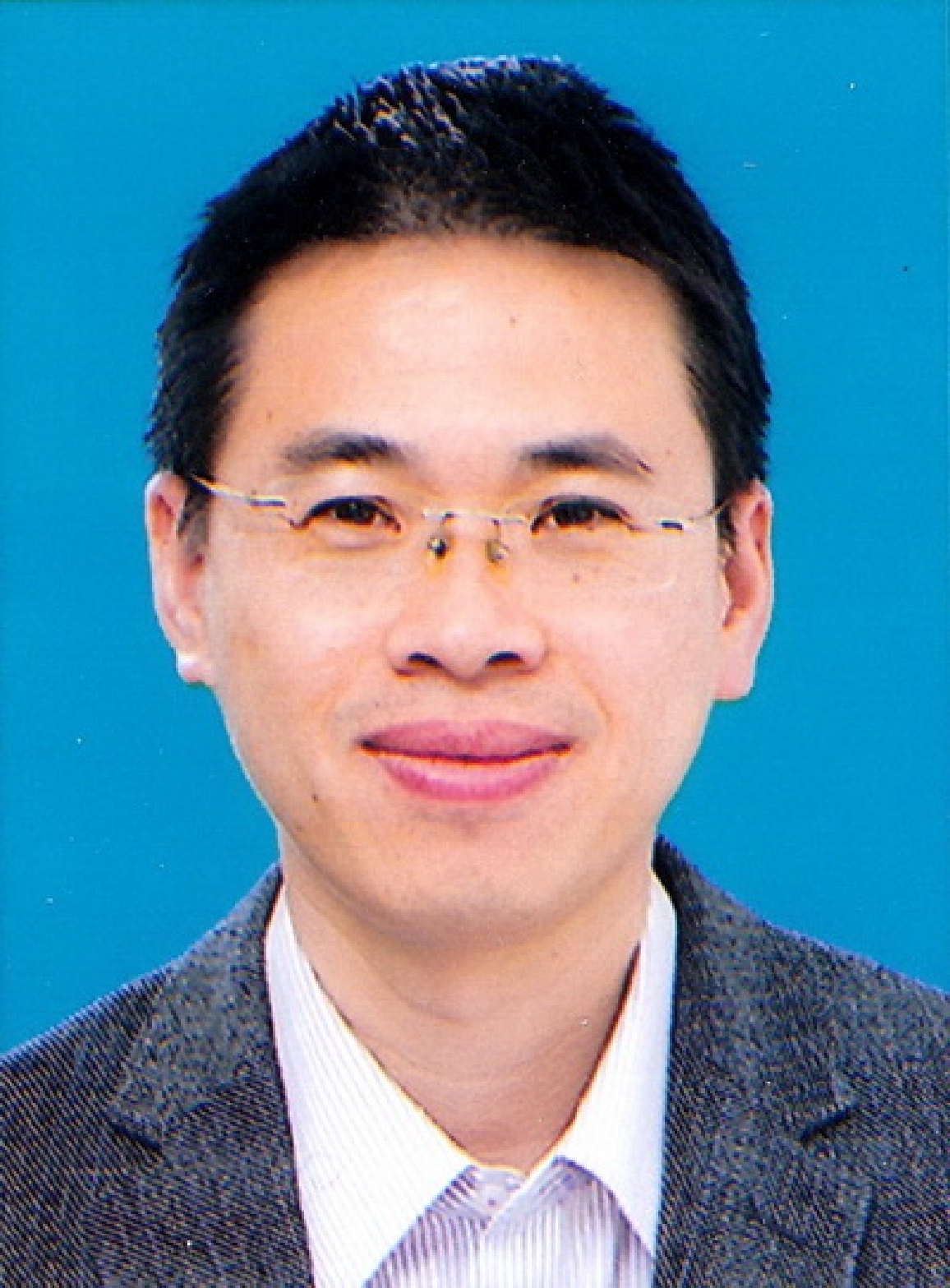}}]{Junzhi Yu} (Fellow, IEEE) received the B.E. degree in safety engineering and the M.E. degree in precision instruments and mechanology from the North University of China, Taiyuan, China, in 1998 and 2001, respectively, and the Ph.D. degree in control theory and control engineering from the Institute of Automation, Chinese Academy of Sciences, Beijing, China, in 2003.

From 2004 to 2006, he was a Postdoctoral Research Fellow with the Center for Systems and Control, Peking University, Beijing. He was an Associate Professor with the Institute of Automation, Chinese Academy of Sciences in 2006, where he became a Full Professor in 2012. In 2018, he joined the College of Engineering, Peking University, as a Tenured Full Professor. His current research interests include intelligent robots, motion control, and intelligent mechatronic systems.
\end{IEEEbiography}

\end{document}